\documentclass[3p,times,procedia]{elsarticle}

\usepackage{lineno,hyperref}
\modulolinenumbers[5]

\usepackage{booktabs} 
\usepackage{multirow}
\usepackage{xcolor}
\newtheorem{myDef}{Definition}
\newtheorem{myExam}{Example}
\usepackage{array}

\usepackage{tabu}
\usepackage{bm}
\usepackage[ruled,linesnumbered,boxed]{algorithm2e}
\usepackage{algorithmic}
\usepackage{subfigure}

\makeatletter
\def\ps@pprintTitle{%
	\let\@oddhead\@empty
	\let\@evenhead\@empty
	\let\@oddfoot\@empty
	\let\@evenfoot\@oddfoot
}
\makeatother

\journal{Artificial Intelligence}

\begin{document}

\begin{frontmatter}

\title{Gradual Machine Learning for Aspect-level Sentiment Analysis}

\author{Yanyan Wang}
\ead{wangyanyan@mail.nwpu.edu.cn}
\author{Qun Chen}
\ead{chenbenben@nwpu.edu.cn}
\author{Jiquan Shen}
\ead{shenjiquan@mail.nwpu.edu.cn}
\author{Boyi Hou}
\ead{ntoskrnl@mail.nwpu.edu.cn}
\author{Murtadha H. M. Ahmed}
\ead{murtadha@mail.nwpu.edu.cn}
\author{Zhanhuai Li}
\ead{lizhh@nwpu.edu.cn}

\address{School of Computer Science, Northwestern Polytechnical University}
\address{Key Laboratory of Big Data Storage and Management, Northwestern Polytechnical University, Ministry of Industry and Information Technology, Xi’an Shaanxi,P.R.China }





%
%
%

\begin{abstract}
The state-of-the-art solutions for Aspect-Level Sentiment Analysis (ALSA) were built on a variety of deep neural networks (DNN), whose efficacy depends on large amounts of accurately labeled training data. Unfortunately, high-quality labeled training data usually require expensive manual work, and may thus not be readily available in real scenarios. In this paper, we propose a novel solution for ALSA based on the recently proposed paradigm of gradual machine learning, which can enable effective machine labeling without the requirement for manual labeling effort. It begins with some easy instances in an ALSA task, which can be automatically labeled by the machine with high accuracy, and then gradually labels the more challenging instances by iterative factor graph inference. In the process of gradual machine learning, the hard instances are gradually labeled in small stages based on the estimated evidential certainty provided by the labeled easier instances. Our extensive experiments on the benchmark datasets have shown that the performance of the proposed solution is considerably better than its unsupervised alternatives, and also highly competitive compared to the state-of-the-art supervised DNN techniques. 
\end{abstract}


\end{frontmatter}

\section{Introduction}

Aspect-level sentiment analysis~\cite{SchoutenF16}, a fine-grained classification task, is highly valuable to both consumers and companies because it can provide the detailed opinions expressed towards certain aspects of an entity. In the literature~\cite{LiX18}, the task of aspect-level sentiment analysis (ALSA) has been classified into two finer subtasks, aspect-term sentiment analysis (ATSA) and aspect-category sentiment analysis (ACSA). ATSA aims to predict the sentiment polarity associated with an explicit aspect term appearing in the text. ACSA instead deals with both explicit and implicit aspects. It needs to predict the sentiment polarities of all the pre-specified aspects in a review, even though an aspect term may not explicitly appear in the text.  
For instance, consider the running example shown in Table~\ref{table:runningexample}, in which $r_i$ and $s_{ij}$ denote the review and sentence identifiers respectively. The review $r_2$ expresses the opinions about a laptop from two aspects, \textit{battery} and \textit{performance}.
The goal of ATSA is to predict the sentiment polarity toward the explicit aspect \textit{battery}; while ACSA has to identify the aspect polarities of both \textit{battery} and \textit{performance} even though the aspect term of \textit{performance} does not appear in the text. In this paper, we target both ATSA and ACSA.

\begin{table}     
	\caption{A running example from laptop reviews.}
	\label{table:running-example}
	\centering
	\begin{tabular}{c|c|c}
		\toprule[1.5pt]
		\textbf{$r_i$}  &
		\textbf{$s_{ij}$}  &
		\textbf{Text} 
		\\ \hline
		\multirow{2}{*}{\begin{tabular}[c]{@{}c@{}} $r_1$ \end{tabular}}
		& $s_{11}$
		& I  {\color{red} \textbf{like}}  the battery that can last {\color{blue} \textbf{long}} time. \\
		& $s_{12}$
		& However, the keyboard sits a little {\color{blue} \textbf{far}} back for me.
		\\ \hline
		\multirow{2}{*}{\begin{tabular}[c]{@{}c@{}} $r_2$ \end{tabular}}
		& $s_{21}$
		& The laptop has a {\color{blue} \textbf{long}} battery life. 
		\\
		& $s_{22}$
		& It also can run my games {\color{blue} \textbf{smoothly}}.
		\\ 
		\bottomrule[1.5pt]		
	\end{tabular}
	
	\label{table:runningexample}
\end{table}


The state-of-the-art techniques for aspect-level sentiment analysis were built on a variety of DNN models~\cite{LiX18,WangHZZ16,RuderGB16}. Compared with previous learning models~\cite{KiritchenkoZCM14, Saias15}, the DNN models can effectively improve classification accuracy by automatically learning multiple levels of representation from data. However, their efficacy depends heavily on large amounts of accurately labeled training data. Unfortunately, high-quality labeled data usually require expensive manual work, and may thus not be readily available in real scenarios. 
To address the limitation resulting from such dependence, this paper presents a novel solution based on the recently proposed paradigm of 
Gradual Machine Learning (GML) ~\cite{HouCSLZWCL19}, which can enable effective machine labeling without the requirement for manual labeling effort. Inspired by the gradual nature of human learning, which is adept at solving problems with increasing hardness, GML begins with some easy instances in a task, which can be automatically labeled by the machine with high accuracy, and then gradually reasons about the labels of the more challenging instances based on the observations provided by the labeled instances.



As pointed out in~\cite{HouCSLZWCL19}, 
even though there already exist many learning paradigms, including transfer learning \cite{PanY10}, lifelong learning \cite{2018Chen}, curriculum learning \cite{BengioLCW09}, and self-training learning~\cite{Mihalcea04} to name a few, GML is fundamentally distinct from them due to its following two properties:
\begin{itemize}	
	\item Distribution misalignment between easy and hard instances in a task. The scenario of gradual machine learning does not satisfy the i.i.d (independent and identically distributed) assumption underlying most existing machine learning models. In GML, the labeled easy instances are not representative of the unlabeled hard instances. The distribution misalignment between the labeled and unlabeled instances renders most existing learning models unfit for gradual machine learning.
	\item Gradual learning by small stages in a task. Gradual machine learning proceeds in small stages. At each stage, it typically labels only one instance based on the evidential certainty provided by the labeled easier instances. The process of iterative labeling can be performed in an unsupervised manner without requiring any human intervention. 
\end{itemize}

We summarize the major contributions of this paper as follows:
\begin{itemize}
	
	\item We propose a novel solution for ALSA based on the paradigm of gradual machine learning, which can enable effective machine labeling without the requirement for manual labeling effort. We have presented the corresponding techniques for the three steps required by the paradigm, which include easy instance labeling, common feature extraction and influence modeling, and scalable gradual inference. 
	
	\item We have empirically evaluated the performance of the proposed solution by a comparative study on benchmark data. Our evaluation results have shown that its performance is considerably better than its unsupervised alternatives, and also highly competitive compared to the state-of-the-art supervised DNN techniques. Moreover, the GML solution is robust in that its performance is, to a large extent, insensitive w.r.t various algorithmic parameters.   
	
\end{itemize}

The rest of this paper is organized as follows: Section~\ref{sec:relatedwork} reviews related work. Section~\ref{sec:preliminaries} defines the task of ALSA and provides a paradigm overview of gradual machine learning. Section~\ref{sec:solution} proposes the technical solution for ALSA. Section~\ref{sec:scalablesolution} presents the scalable solution of gradual inference for ALSA. Section~\ref{sec:experiments} empirically evaluates the performance of the proposed solution. Finally, we conclude this paper with Section~\ref{sec:conclusion}.

\section{Related work} \label{sec:relatedwork}


\vspace{0.05in}
\hspace{-0.20in}{\bf Machine Learning Paradigms.} We first proposed the paradigm of gradual machine learning and applied it on the task of entity resolution in~\cite{HouCSLZWCL19,Houarxiv}. There exist many other machine learning paradigms proposed for a wide variety of classification tasks. Here we will briefly review those closely related to GML and discuss their difference from GML.

Traditional machine learning algorithms make predictions on the future data using the statistical models that are trained on previously collected labeled or unlabeled training data~\cite{Christen08, MullenC04, BlumM98, BellareIPR12}. In many real scenarios, the labeled data may be too few to build a good classifier. Semi-supervised learning~\cite{NigamMTM00, XiangZ14, HeS17} addresses this problem by making use of a large amount of unlabeled data and a small amount of labeled data. Nevertheless, the efficacy of both supervised and semi-supervised learning paradigms depends on the i.i.d assumption. Therefore, they can not be applied to the scenario of gradual machine learning.

In contrast, transfer learning~\cite{PanY10}, allows the distributions of the data used in training and testing to be different. It focused on using the labeled training data in a domain to help learning in another target domain. The other learning techniques closely related to transfer learning include lifelong learning~\cite{2018Chen} and multi-task learning~\cite{Caruana97}. Lifelong learning is similar to transfer learning in that it also focused on leveraging the experience gained on the past tasks for the current task. However, different from transfer learning, it usually assumes that the current task has good training data, and aims to further improve the learning using both the target domain training data and the knowledge gained in past learning. Multi-task learning instead tries to learn multiple tasks simultaneously even when they are different. A typical approach for multi-task learning is to uncover the pivot features shared among multiple tasks. However, all these learning paradigms can not be applied to the scenario of gradual machine learning. Firstly, focusing on unsupervised learning within a task, gradual machine learning does not enjoy the access to good labeled training data or a well-trained classifier to kick-start learning. Secondly, the existing techniques transfer instances or knowledge between tasks in a batch manner. As a result, they do not support gradual learning by small stages within a task.


The other related machine learning paradigms include curriculum learning (CL)~\cite{BengioLCW09} and self-paced learning (SPL)~\cite{KumarPK10}. Both of them are, to some extent, similar to gradual machine learning in that they were also inspired by the learning principle underlying the cognitive process in humans, which generally start with learning easier aspects of a task, and then gradually takes more complex examples into consideration. However, both of them depend on a curriculum, which is a sequence of training samples essentially corresponding to a list of samples ranked in ascending order of learning difficulty. A major disparity between them lies in the derivation of the curriculum. In CL, the curriculum is assumed to be given by an oracle beforehand, and remains fixed thereafter. In SPL, the curriculum is instead dynamically generated by the learner itself, according to what the learner has already learned. Based on the traditional learning models, both CL and SPL depend on the i.i.d assumption and require good-coverage training examples for their efficacy. However, the scenario of gradual machine learning does not satisfy the i.i.d assumption. GML actually aims to eliminate the dependency on good-coverage training data.

Online learning~\cite{KivinenSW01} and incremental learning~\cite{SchlimmerG86} have also been proposed for the scenarios where training data only becomes available gradually over time or its size is out of system memory limit. Based on the traditional learning models, both online learning and incremental learning depend on high-quality training data for their efficacy. Therefore, they can not be applied for gradual learning either. 

\vspace{0.15in}
\hspace{-0.20in}{\bf Work on Aspect-level Sentiment Analysis.}
Sentiment analysis at different granularity levels, including document, sentence, and aspect levels, has been extensively studied in the literature \cite{KumarR15}. The early work for aspect-level sentiment analysis~\cite{KiritchenkoZCM14,Saias15} focused on traditional machine learning algorithms. Unfortunately, the performance of these traditional techniques depends heavily on the quality of the features, which usually have to be manually crafted.


Since deep neural networks can automatically learn high-quality features or representations, most recent work attempted to adapt such models for aspect-level sentiment analysis. The existing work can be divided into two categories based on the two finer subtasks of ATSA and ACSA. 
For ATSA task, Dong \cite{DongWTTZX14} initially proposed an Adaptive Recursive Neural Network (AdaRNN) that can employ a novel multi-compositionality layer to propagate the sentiments of words towards the target. Noticing that the models based on recursive neural network heavily rely on external syntactic parser, which may result in inferior performance, many works subsequently focused on recurrent neural networks. Tang \cite{TangQFL16} developed a target-dependent LSTM (TD-LSTM) model to capture the connection between target words and their contexts. The alternative solutions include memory networks and convolutional neural networks. Wang~\cite{LiuCWMZ18} proposed  
Target-sensitive Memory Networks that can capture the sentiment interaction between targets and contexts. Li~\cite{LamLSB18} presented Transformation Networks that employ a CNN layer to extract salient features from the transformed word representations originated
from a bi-directional RNN layer. Due to the great success of attention mechanism in image recognition \cite{MnihHGK14}, speech recognition \cite{ChorowskiBSCB15},  machine translation \cite{LuongPM15, FiratCB16} and question answering \cite{HeG16},
many models based on LSTM and attention mechanism have also been proposed. These models, including Hierarchical Attention Network~\cite{ChengZZKZW17}, Segmentation Attention Network~\cite{WangL18a}, Interactive Attention Networks~\cite{MaLZW17}, Recurrent Attention Network~\cite{ChenSBY17}, Attention-over-Attention Neural Networks~\cite{HuangOC18}, Effective Attention Modeling~\cite{HeLND18}, Content Attention Model~\cite{LiuZZHW18}, Multi-grained Attention Network~\cite{FanFZ18}, employed different attention mechanisms to output the aspect-specific sentiment features.

In comparison, there exist fewer works for ACSA because the implicit aspects make the task more challenging. Ruder~\cite{RuderGB16} proposed a hierarchical bidirectional LSTM for the ACSA task by modeling the inter-dependencies of sentences in a review.
Wang~\cite{WangHZZ16} presented an attention-based LSTM that employs an aspect-to-sentence attention mechanism to concentrate on the key part of a sentence given an aspect. 
Xue~\cite{LiX18} introduced a model based on convolutional neural networks and gating mechanisms, which has been empirically shown to be
more accurate and efficient compared to its previous alternatives. It is noteworthy that the DNN models proposed for ACSA can also be used for ATSA, but the models proposed for ATSA are usually not applicable to ACSA because they employ specific mechanisms
to model an explicit aspect term along with its relative context.

However, the efficacy of the aforementioned DNN-based approaches depends heavily on good-coverage training data, which may not be readily available in real scenarios.

\section{Preliminaries}  \label{sec:preliminaries}
In this section, we first define the task of aspect-level sentiment analysis, and then provide a paradigm overview of gradual machine learning.

\subsection{Task Statement}

\begin{table}
	\caption{Frequently used notations.}
	\centering
	\begin{tabular}{l|l}
		\hline
		\toprule[1.5pt]
		Notation & Description \\ \hline			
		$r_j$ & a review \\ 
		$s_k$ & a sentence \\ 
		$a_l$ & an aspect category or aspect term \\ 
		$t_i=(r_j,s_k,a_l)$ & an aspect unit \\ 
		$T=\{{t_i}\}$ & a set of aspect units \\ 
		$v_i$ & a boolean variable indicating the polarity of the aspect unit $t_i$ \\
		$V=\{v_i\}$ & a set of aspect polarity variables \\

		\bottomrule[1.5pt]				
	\end{tabular}
	\label{table:frequently_notations}
\end{table}

For presentation simplicity, we have summarized the frequently used
notations in Table~\ref{table:frequently_notations}. Formally, we formulate the task of aspect-level sentiment analysis as follows:

\begin{myDef}
	\emph{\textbf{[Aspect-level Sentiment Analysis]}} Let $t_i=(r_j,s_k,a_l)$ be an aspect unit, where $r_j$ denotes a review, $s_k$ denotes a sentence in the review $r_j$, and $a_l$ denotes an aspect associated with the sentence $s_k$. Note that the aspect $a_l$ can be a aspect category or aspect term, and a sentence may express opinions towards multiple aspects. Given a corpus of reviews, $R$, the goal of the task is to predict the sentiment polarity of each aspect unit $t_i$ in $R$. 
\end{myDef}

In this paper, we suppose that an aspect polarity is either positive or negative. 

\subsection{Paradigm overview} \label{sec:paradigm}

\begin{figure*}
	\centering
	\includegraphics[scale=0.26]{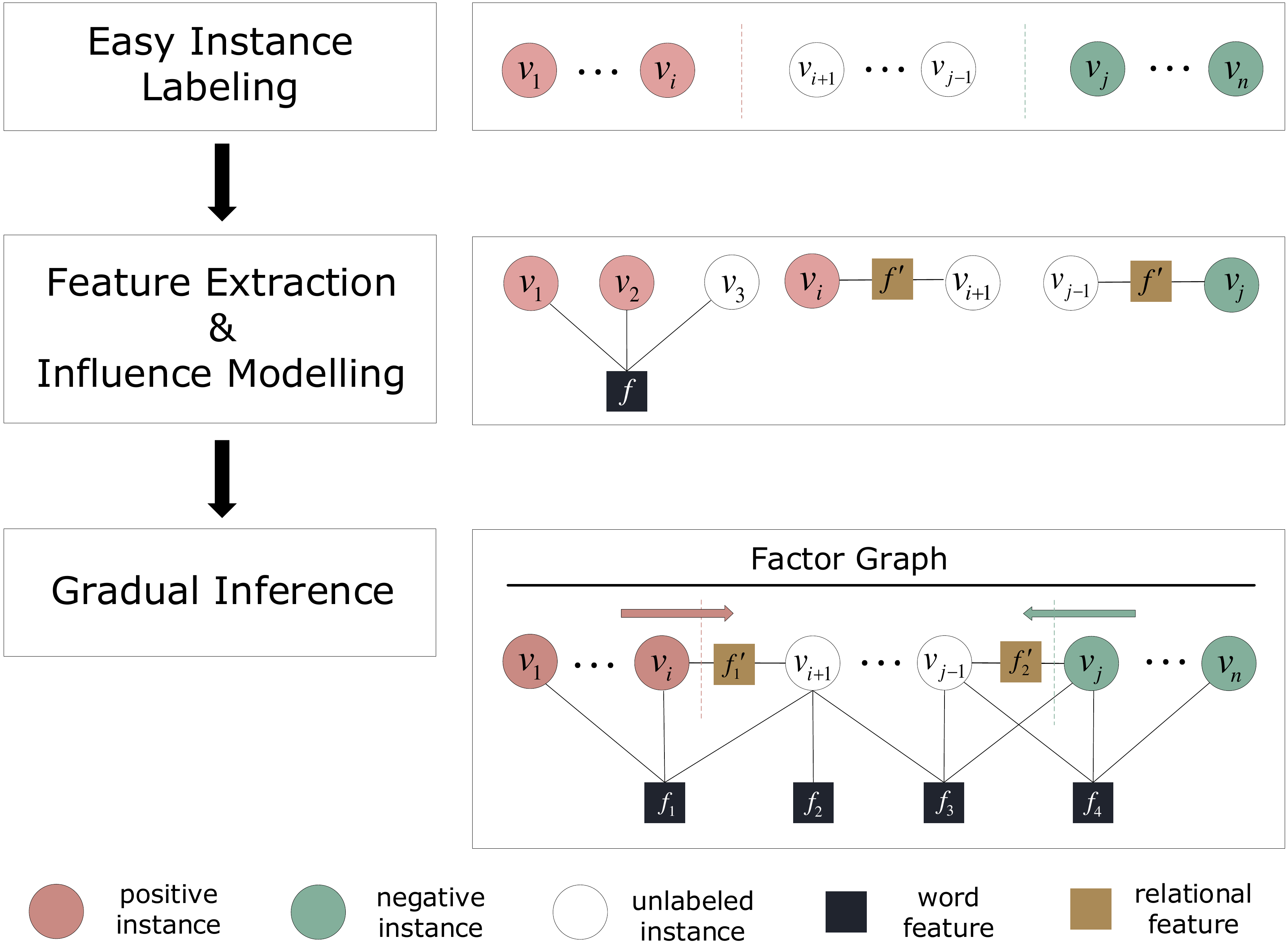}
	\caption{Learning paradigm overview.}
	\label{fig:framework}
\end{figure*}

The process of gradual machine learning, as shown in Figure~\ref{fig:framework}, consists of the following three steps:

\begin{itemize}
	\item {\bf Easy Instance Labeling.} Given a classification task, it is usually very challenging to accurately label all the instances in the task without good-coverage training examples. However, the work can become much easier if we only need to automatically label some easy instances in the task. In real scenarios, easy instance labeling can be performed based on the simple user-specified rules or the existing unsupervised learning techniques. For instance, in unsupervised clustering, an instance close to the center of a cluster in the feature space
	can usually be considered as an easy instance, because it only has a remote chance to be misclassified. Gradual machine learning begins with the observations provided by the labels of easy instances. Therefore, high accuracy of automatic machine labeling on easy instances is critical for its ultimate performance on a given task.
	
	\item {\bf Feature Extraction and Influence Modeling.} Feature serves as the medium to convey the knowledge obtained from the labeled easy instances to the unlabeled harder ones. This step extracts the common features shared by the labeled and unlabeled instances. To facilitate effective knowledge conveyance, it is desirable that a wide variety of features are extracted to capture as much information as possible. For each extracted feature, this step also needs to model its influence over the labels of its relevant instances.

	\item {\bf Gradual Inference.} This step gradually labels the instances with increasing hardness in a task. Since the scenario of gradual learning does not satisfy the i.i.d assumption, gradual learning is fulfilled from the perspective of evidential certainty. As shown in Figure~\ref{fig:framework}, it constructs a factor graph, which consists of the labeled and unlabeled instances and their common features. Gradual learning is conducted over the factor graph by iterative inference. At each iteration, it chooses to label the unlabeled instance with the highest degree of evidential certainty. The iteration is repeatedly invoked until all the instances in a task are labeled. In gradual inference, a newly labeled instance at the current iteration would serve as an evidence observation in the following iterations.
	
\end{itemize}

\section{Solution for ALSA}  \label{sec:solution}


This section proposes the solution of gradual machine learning for ALSA. We will present the corresponding techniques for the three steps laid out in Subsection~\ref{sec:paradigm}.

\subsection{Easy Instance Labeling}

The existing lexicon-based approaches~\cite{DingLY08} essentially reason about polarity by summing up the polarity scores of all the sentiment words in a sentence. The score of a sentiment word indicates its intensity of sentiment, which increases with the absolute value of score. Since negation words can effectively reverse the polarity of a sentiment word, they usually perform negation detection for each sentiment word by examining whether there is any negation in its neighboring words \cite{HuttoG14}.

Unfortunately, the lexicon-based approaches are prone to error under some ambiguous circumstances. Firstly, the presence of contrast (e.g. \textit{but} and \textit{although}), hypothetical (e.g. \textit{if}) or condition (e.g. \textit{unless}) connectives could significantly complicate polarity detection. For instance, the sentence, ``would be a very nice laptop if the mousepad worked properly'', contains only the positive sentiment words ``nice'' and ``properly'', but it holds negative attitude due to the presence of the hypothetical connective ``if''. Secondly, the negation words involving long-distance dependency could also make polarity detection challenging. For instance, in the sentence, ``I don't really think the laptop has a good battery life'', the negation word ``don't'' reverses the polarity, but it is far away from the sentiment word ``good''. Finally, a sentence may contain multiple sentiment words that hold conflicting polarities; in this case, its true polarity is not easily detectable based on sentiment word scoring. 


Therefore, we identify easy instances by excluding the aforementioned ambiguous circumstances as follows:

\begin{myDef}
	\emph{\textbf{[Easy Instance]}} 
	We consider an aspect polarity, $t_i=(r_j,s_k,a_l)$, as an easy instance if and only if the sentence expressing opinions about the aspect, $s_k$, simultaneously satisfies the following three conditions:
	\begin{itemize}
		\item It contains at least one sentiment word, but does not simultaneously contain any sentiment word holding a conflicting polarity;
		\item It does not contain any contrast, hypothetical or condition connective;
		\item It does not contain any negation word involving long-distance dependency.
	\end{itemize}
	\label{def:easy}
\end{myDef}

The polarity of an easy instance is simply determined by the polarity of its sentiment words.
Moreover, a negation word is supposed to involve long-distance dependency if and only if it is not in the neighboring 3-grams preceding any sentiment word. We illustrate the difference between the easy and challenging instances by Example~\ref{exam:easy}.

\begin{myExam}
	\emph{\textbf{[Easy Instances]}}
	In a phone review, the sentence, ``the screen is not good for carrying around in your bare hands'', which expresses opinion about ``screen'', is an easy instance because the sentiment word ``good'' associated with the local negation cue ``not'' strongly indicates the negative sentiment.
	In contrast, the sentence, ``I don't know why anyone would want to write a great review about this battery'',  which expresses opinion about ``battery'', is not an easy instance. Even though it contains the sentiment word ``great'', it also includes the negation word ``don't'' involving long-distance dependency. Similarly, the sentence, ``I like this laptop, the only problem is that it can not last long time'', is not an easy instance because it contains both positive and negative words, i.e. ``like'' and ``problem'' respectively. 
	\label{exam:easy}
\end{myExam}

\subsection{Feature Extraction and Influence Modeling}
\label{subsec:feature-extraction}

We extract two types of features for influence modeling: word feature and relational feature.


\vspace{0.05in}
\hspace{-0.20in}{\bf Word Feature.}  Sentiment polarity is usually determined by sentiment words. Therefore, we extract sentiment words, which have been specified in the open-source lexicons, from sentences and consider them as the features of aspect polarities. To capture more information shared among aspect instances, besides the single sentiment words, we also extract k-grams ($k\geq 2$) as word features. Since negation word can effectively reverse the polarity of a sentiment word~\cite{JiaYM09, HogenboomIHFK11}, we perform negation detection for each sentiment word by examining whether there is any negation in its neighboring words.

Note that a sentence may express opinions towards multiple aspects. In the case of ATSA, it is easy to correlate a word feature in a sentence with its target aspect because the aspect term explicitly appears in the text. In the case of ACSA, we first employ the dependency-based parse tree to extract all the pairs of opinion target and opinion word~\cite{Abbasi2013Aspect}, and then assign an opinion word to an aspect if either itself or its opinion target is close to the aspect in the vector space (namely, their similarity exceeds a threshold (e.g. 0.5 in our implementation)). For k-gram feature extraction, we first extract the opinion phrase towards an aspect based on its opinion words, and then assign the k-gram features contained in the opinion phrase to the aspect.



\vspace{0.05in}
\hspace{-0.20in}{\bf Relational Feature.} Modeling sentences independently, the existing DNN models for aspect-level sentiment analysis have very limited capability in capturing the contextual information at sentence level. However, the sentences in a review build upon each other. There often exist some discourse relations between clauses or sentences, which can provide valuable hints for polarity reasoning.  Specifically, it can be observed that two sentences connected with a shift word  usually have opposite polarities. In contrast, two neighboring sentences without any shift word between them usually have similar polarities. In the running example shown in Table~\ref{table:running-example}, the polarities of $s_{11}$ and $s_{12}$ are opposite because they are connected by the shift word of ``but'', while the polarities of $s_{21}$ and $s_{22}$ are similar due to the absence of any shift word between them. 


Therefore, we use the rules to extract the similar and opposite relations between aspect polarity based on sentence context. Given two aspect units $t_i=\{r_i, s_i, a_i\}$ and $t_j=\{r_j,s_j, a_j\}$ that are opinioned in the same review (namely $r_i=r_j$), the rules for polarity relation extraction are specified as follows:
\begin{enumerate}
	\item If the sentences $s_i$ and $s_j$ are identical ($s_i$=$s_j$) or adjacent and neither of them contains any shift word, $t_i$ and $t_j$ are supposed to hold similar polarities;
	\item If two adjacent sentences $s_i$ and $s_j$ are connected by a shift word and neither of them contains any inner-sentence shift word, $t_i$ and $t_j$ are supposed to hold opposite polarities;
	\item If the sentences $s_i$ and $s_j$ are identical and the opinion clauses associated with them are connected by an inner-sentence shift word, $t_i$ and $t_j$ are supposed to hold opposite polarities. 
\end{enumerate}

Given an ATSA task, it is easy to correlate an opinion clause with its target aspect because the aspect term explicitly appears in the text. Therefore, the condition specified in the 3rd rule can be easily checked in the scenario of ATSA. The scenario of ACSA is instead more challenging. Our solution first uses the dependency-based parse tree to extract all the opinion phrases, and associates an opinion clause with a specific aspect if either its opinion target or opinion word is close to the aspect in the vector space.

\subsection{Gradual Inference}

\begin{figure}
	\centering
	\includegraphics[scale=0.48]{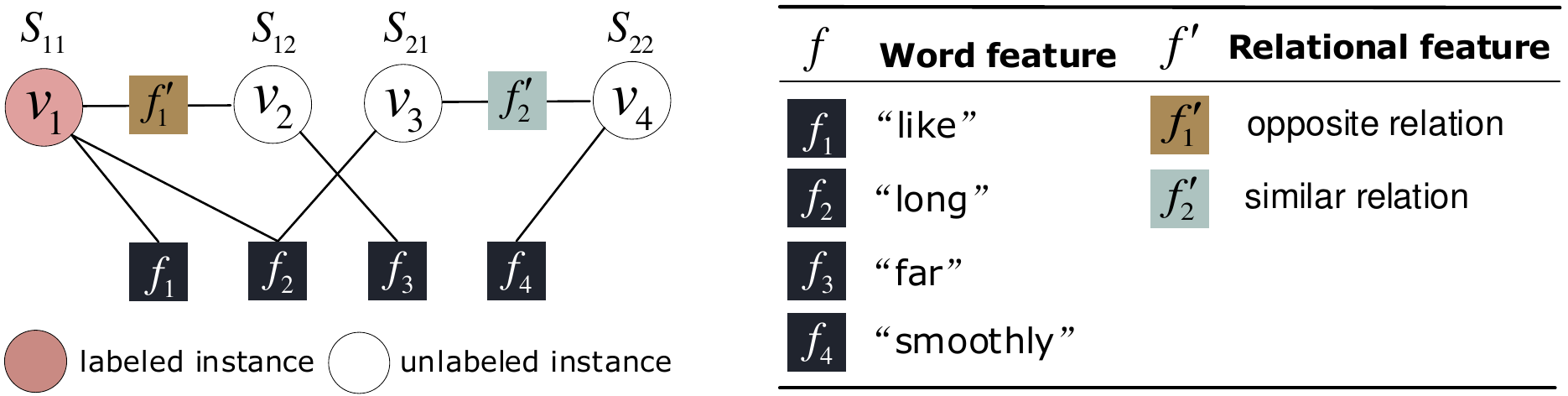}
	\caption{The factor graph constructed for the running example. }
	\label{fig:factorgraph}
\end{figure}


As usual, we construct a factor graph, $G$, in which the labeled easy instances are represented by the \emph{evidence variables}, the unlabeled hard instances by the \emph{inference variables}, and the features by the \emph{factors}. The value of each variable represents its corresponding polarity. An evidence variable has the constant value of 0 or 1, which indicate the polarity of \emph{negative} and \emph{positive} respectively. The values of evidence variables remain unchanged during the inference process. The values of the inference variables should instead be inferred based on $G$. The factor graph constructed for the running example has been shown in Figure~\ref{fig:factorgraph}.

Gradual machine learning is attained by iterative factor graph inference on $G$. In $G$, we define the probability distribution over its variables $V$ by
\begin{equation}
\label{eq:fg model}
P_w(V)=\frac{1}{Z_w}\prod_{v \in V} \prod_{f \in F_v} \phi_{f}(v)
\prod_{f' \in F'}\phi_{f'}(v_i, v_j),
\end{equation}
where $F_v$ denotes the set of word features associated with the variable $v$, $F'$ denotes the set of relational features, $\phi_{f}(v)$ denotes the factor associated with $v$ and $f$, and $\phi_{f'}(v_i, v_j)$ denotes the factor associated with the relational feature $f'$. In Eq.~\ref{eq:fg model}, the factor of a word feature $f$ is defined by
\begin{equation}
\phi_{f}(v) = 
\left \{
\begin{array}{ll}
1 & v=0; \\
e^{w_f} & v=1;  
\end{array} 
\right.
\end{equation}
where $v$ denotes a variable having the feature $f$, and $w_f$ denotes the weight of $f$. Similarly, the factor of a relational feature $f'$ is defined by 
\begin{equation}
\label{eq:binary factor}
\phi_{f'}(v_i, v_j) = 
\left \{
\begin{array}{ll}
e^{w_{f'}}     &      if \ v_i = v_j; \\
1    &   otherwise;  
\end{array} 
\right.
\end{equation}
where $v_i$ and $v_j$ denote the two variables sharing the feature $f'$, and $w_{f'}$ denotes the weight of $f'$. 
Note that the weight of a word factor can be positive or negative, while the weight of a \emph{similar} relational factor is positive and the weight of an \emph{opposite} relational factor is negative. In our implementation, the weights of all the \emph{similar} relational factors are set to be the same; the weights of all the \emph{opposite} relational factors are also set to be the same.

As in \cite{HouCSLZWCL19}, 
given a factor graph with some labeled evidence variables, we reason about the factor weights by minimizing the negative log marginal likelihood of
\begin{equation} \label{eq:weight-learning}
\hat w  = arg \min \limits_{w} -log \sum_{V_I} P_w(\Lambda, V_I),
\end{equation}
where $\Lambda$ denotes the observed labels of evidence variables and $V_I$ denotes the set of inference variables. The objective function effectively learns the factor weights most consistent with the label observations of evidence variables. 




As usual, gradual inference proceeds in small stages. At each stage, it chooses to label the unlabeled variable with the highest degree of evidential certainty in $G$. The iteration is repeatedly invoked until all the inference variables are labeled. In gradual inference, evidential certainty is measured by the inverse of entropy. Formally, entropy is formally defined by
\begin{equation}
H(v)=-(P(v)\cdot {\ln}P(v) + (1-P(v))\cdot {\ln}(1-P(v))),
\end{equation}
in which $H(v)$ denotes the entropy of a variable $v$. 

In our implementation, we have used the Numbskull library \footnote{https://github.com/HazyResearch/numbskull} to optimize this objective function by interleaving stochastic gradient descent steps with Gibbs sampling ones, similar to contrastive divergence. 
However, repeated inference by maximum likelihood over a large-sized factor graph is usually very time-consuming~\cite{zhou2016archimedesone}. Therefore, in the next section, we will present a scalable solution for gradual inference on ALSA.



\section{Scalable Gradual Inference} \label{sec:scalablesolution}

We have built the scalable solution based on the framework proposed in \cite{Houarxiv}, which consists of three steps, measurement of evidential support, approximate ranking of entropy and construction of inference subgraph.  Its efficacy depends on the following observations:
\begin{itemize}
	\item Many unlabeled inference variables in the factor graph are only weakly linked through the factors to the evidence variables. Due to the lack of evidential support, their inference probabilities would be quite ambiguous, i.e. close to 0.5. As a result, only the inference variables that receive considerable support from the evidence variables need to be considered for labeling;
	\item With regard to the probability inference of a single variable $v$ in a large factor graph, it can be effectively approximated by considering the potentially much smaller subgraph consisting of $v$ and its neighboring variables. The inference over the subgraph can usually be much more efficient than over the original entire graph.
\end{itemize}

\begin{algorithm}[t]
	\caption{Scalable Gradual Inference}
	\label{alg:gradualinference}
	\While{there exists any unlabeled variable in $G$}
	{
		$V' \leftarrow$ all the unlabeled variables in $G$\;
		\For{$v\in V'$}
		{
			Measure the evidential support of $v$ in $G$\;
		}
		Select top-$m$ unlabeled variables with the most evidential support (denoted by $V_m$) \;
		\For{$v\in V_m$}
		{
			Approximately rank the entropy of $v$ in $V_m$\;
		}
		Select top-$k$ most promising variables in terms of entropy in $V_m$ (denoted by $V_k$) \;
		\For{$v\in V_k$}
		{
			Compute the probability of $v$ in $G$ by factor graph inference over a subgraph of $G$\;
		}
		Label the variable with the minimal entropy in $V_k$\;
	}
\end{algorithm}

The process of scalable gradual inference is sketched in Algorithm~\ref{alg:gradualinference}. Given a factor graph $G$, it first selects the top-$m$ unlabeled variables with the most evidential support in $G$ as the candidates for probability inference. To reduce the invocation frequency of factor graph inference, it then approximates entropy estimation by an efficient algorithm on the $m$ candidates and selects only the top-$k$ most promising variables among them for factor graph inference. Finally, it infers the probabilities of the chosen $k$ variables in $G$. For each variable, its probability is not inferred over the entire graph of $G$, but over a potentially much smaller subgraph.


\subsection{Measurement of Evidential Support} \label{sec:evidentialsupport}

Given an inference variable $v$, we first estimate the evidential support provided by each of its factors, and then aggregate them to measure its overall evidential support based on the Dempster-Shafer (D-S) theory~\cite{YangX13}.


Formally, let $X$ be the universal set representing all possible states of a system under consideration. By a function of basic belief assignment, the D-S theory assigns a belief mass to each element of the power set of $X$. The mass of an element $E$, denoted by $m(E)$, expresses the proportion of all relevant and available evidence that supports the claim that the actual state belongs to $E$ but to no particular subset of $E$. The masses of elements satisfy
\begin{equation} 
\sum_{E\in 2^X}{m(E)}  =1 \hspace{0.05in} \& \hspace{0.05in} m(\emptyset)=0.
\end{equation}
In case that only singleton propositions are assigned belief masses, a mass function reduces to a classical probability function.


For evidential support measurement, we define two propositions: ``label the instance", denoted by $L$, and ``unlabel the instance, denoted by $U$. With $X=\{L, U\}$, the power set of $X$ can be represented by $2^X=\{\emptyset, L, U, X\}$. Given an inference variable $v$ and its word feature $f$, we estimate the evidential support that $v$ receives from $f$ by the mass function
\begin{equation} \label{eq:word-mass}
m_{f}(E) = 
\left \{
\begin{array}{ll}
(1-d_{f})\cdot max\{P(f), 1-P(f)\} & E=\{L\}, \\
(1-d_{f})\cdot min\{P(f), 1-P(f)\} & E=\{U\}, \\
d_{f}     &   E=\{L, U\},  
\end{array} 
\right.
\end{equation}
where $d_{f}$ denotes the degree of uncertainty of $f$, and $P(f)$ denotes the proportion of positive instances among all labeled instances having the feature $f$. According to Eq.~\ref{eq:word-mass}, the mass assigned to the element of \{$L$\} increases as the value of $P(f)$ becomes more extreme (i.e. close to 0 or 1). The underlying intuition is that the more extreme the value of $P(f)$ is, the more evidential support the element of \{$L$\} should receive from the feature $f$.


Similarly, given an inference variable $v$ and its relational feature $f'$, we estimate the evidential support that $v$ receives from $f'$ by the mass function
\begin{equation} \label{eq:relation-mass}
m_{f'}(E) = 
\left \{
\begin{array}{ll}
(1-d_{f'})\cdot R(f')    &      E = \{L\}, \\
(1-d_{f'})\cdot (1-R(f'))    &   E = \{U\}, \\
d_{f'}     &   E = \{L, U\},  
\end{array} 
\right.
\end{equation}
where $d_{f'}$ denotes the degree of uncertainty of $f'$, and $R(f')$ denotes the accuracy of the relation $f'$. 
In Eq.~\ref{eq:relation-mass}, $R(f')$ can be considered as the statistical accuracy of the extracted relations, which can be estimated based on labeled instances; the evidential support that the element of \{$L$\} receives from $f'$ thus increases with the estimated accuracy.


We are now ready to describe how to measure the aggregate evidential support provided by multiple factors. Suppose that an inference variable $v$ has $i$ word features, \{$f_1$,$\ldots$,$f_i$\}, and $j$ relational features, \{$f_1'$,$\ldots$,$f_j'$\}. Given the element of $E=\{L\}$, we estimate its aggregate evident support by combining the estimated masses as follows
\begin{equation}
m(E)=m_{f_1}(E)\oplus \cdots \oplus m_{f_i}(E) \oplus m_{f'_1}(E) \oplus \cdots \oplus m_{f'_j}(E),
\end{equation}
where $m(E)$ denotes the total amount of evidential support that $v$ receives, and the combination is calculated from the two sets of mass functions,  $m_{f_1}(E)$ and $m_{f_2}(E)$, as follows
\begin{equation}  \label{eq:two masses}
m_{f_1}(E)\oplus m_{f_2}(E)={\frac {1}{1-K}}\sum _{E'\cap E''=E}m_{f_1}(E')\cdot m_{f_2}(E''),
\end{equation}
where $E'$ and $E''$ denote the elements of the power set, and
\begin{equation} \label{eq:conflict-measurement}
K=\sum _{E'\cap E''=\emptyset }m_{f_1}(E')\cdot m_{f_2}(E''), 
\end{equation}
which is a measure of the amount of conflict between $E'$ and $E''$. 

Note that the degree of uncertainty, denoted by $d_{f}$ and $d_{f'}$ in Eq.~\ref{eq:word-mass} and ~\ref{eq:relation-mass}, indicates how much impact a feature has on the whole degree of belief in terms of evidential support measurement. The lower the value, the greater the impact. It can be observed that relational features usually provide more reliable information than word features. Therefore, in practical implementation, we suggest that $d_{f'}$ is set to be smaller than $d_{f}$ (e.g., $d_{f}=0.4$ and $d_{f'}=0.1$). Our empirical evaluation in Subsection~\ref{sec:sensitivity} has shown that the performance of gradual machine learning is, to a large extent, insensitive to the parameter setting of $d_{f}$ and $d_{f'}$.

\subsection{Approximate Ranking of Entropy} \label{sec:approximateinference}

Since more evidential conflict means more status uncertainty, we approximate the entropy ranking of inference variables by measuring their evidential conflict. Similar to the case of evidential support measurement, we define two propositions: ``label it as \emph{positive}'', denoted by $L^+$, and ``label it as \emph{negative}'', denoted by $L^-$. Given an inference variable $v$ and its word feature $f$, we approximate $v$'s evidential certainty w.r.t $f$ with the mass function
\begin{equation} \label{eq:word-rank}
m_{f}^{*}(E) = 
\left \{
\begin{array}{ll}
(1-d_{f}^{*})\cdot P(f)    &      E = \{L^+\}, \\
(1-d_{f}^{*})\cdot (1-P(f))    &   E = \{L^-\}, \\
d_{f}^{*}     &   E = \{L^+, L^-\},  
\end{array} 
\right.
\end{equation}
where $d_{f}^{*}$ denotes the degree of uncertainty of $f$, and $P(f)$ denotes the proportion of positive instances among all the labeled instances having the feature $f$. 


Similarly, given an inference variable $v$ and its relational feature $f'$, we approximate $v$'s evidential certainty w.r.t $f'$ with the mass function
\begin{equation} \label{eq:relation-rank}
m_{f'}^{*}(E) = 
\left \{
\begin{array}{ll}
(1-d_{f'}^{*})\cdot P(f')    &      E = \{L^+\},\\
(1-d_{f'}^{*})\cdot (1-P(f'))    &   E = \{L^-\}, \\
d_{f'}^{*}     &   E = \{L^+, L^-\},  
\end{array} 
\right.
\end{equation} 
where $d_{f'}^{*}$ denotes the degree of uncertainty of $f'$, and $P(f')$ denotes the probability of $v$ being positive if only the evidence $f'$ is considered for labeling $v$. If the labeled variable on the other side of the relation $f'$ is positive, we set $P(f')=\frac{e^{w_{f'}}}{1+e^{w_{f'}}}$, in which $w_{f'}$ denotes the weight of $f'$; otherwise (i.e., it is negative), we set $P(f')=\frac{1}{1+e^{w_{f'}}}$.





Finally, we measure the amount of conflict between the multiple pieces of evidence using the generalized expression of $K$ as specified in Eq.~\ref{eq:conflict-measurement}. Similar to the case of evidential support measurement, we suggest that $d_{f'}^{*}$ is set to be a lower value than $d_{f}^{*}$. Our empirical evaluation in Subsection~\ref{sec:sensitivity} has shown that the performance of gradual machine learning is, to a large extent, insensitive to the parameter setting of $d_{f}^{*}$ and $d_{f'}^{*}$.



\subsection{Construction of Inference Subgraph} \label{sec:subgraphconstruction}


It has been empirically shown~\cite{LiZWGDD17} that given a variable $v$ in $G$,  its probability inference can be effectively approximated by considering the subgraph consisting of $v$ and its $r$-hop neighboring variables, and even with a small value of $r$ (e.g. 2 and 3), the approximation can be sufficiently accurate in many real scenarios. Therefore, given a target inference variable $v$ in $G$, we extract all its $2$-hop neighbors reachable by the relational factors and include them in the subgraph. For each word feature of $v$, all the labeled and unlabeled instances sharing the feature with $v$ are also included in the constructed subgraph because potentially, their labels can significantly influence the label of $v$.  






\section{Empirical Evaluation} \label{sec:experiments}


In this section, we empirically evaluate the performance of the proposed solution by comparative study. We have compared GML with the state-of-the-art techniques proposed for both ACSA and ATSA. Note that the DNN models proposed for ACSA can also be used
for ATSA, but the models proposed for ATSA are usually not applicable to ACSA because they employ specific mechanisms
to model an explicit aspect-term along with its relative context.

For the ACSA task, the compared techniques include:
\begin{itemize}
	\item \textbf{LEX-SYN~\cite{Alvarez-LopezJG16}.} It is an unsupervised approach built on lexicons and syntactic dependency analysis;
	
	\item \textbf{VADER~\cite{HuttoG14}.} It is a rule-based method proposed for sentence-level sentiment analysis. We have adapted it for the task of ALSA. Given a sentence with multiple aspects, the solution identifies the sentiment polarity of an aspect by analyzing its opinioned clause, whose extraction has been explained in Subsection~\ref{subsec:feature-extraction}.
	
	
	\item \textbf{H-LSTM~\cite{RuderGB16}.} It is an enhanced DNN model. It models the inter-dependencies of sentences in a review using a hierarchical bidirectional LSTM;
	
	\item \textbf{AT-LSTM~\cite{WangHZZ16}.} Referring to the Attention-based LSTM, it employs an attention mechanism to concentrate on the key parts of a sentence given an aspect, where the aspect embeddings are used to determine the attention weights;
	
	\item \textbf{ATAE-LSTM~\cite{WangHZZ16}.} Referring to the Attention-based LSTM with Aspect Embedding, it is supposed to be an improvement over AT-LSTM. It extends AT-LSTM by appending the input aspect embedding to each word's input vector; 
	
	\item \textbf{GCAE~\cite{LiX18}.} It uses convolutional neural networks and gating mechanisms to predict the sentiment polarity of a given aspect. Compared with the LSTM and attention mechanisms, it can be more accurate and efficient.
	
\end{itemize}


For the ATSA task, besides LEX-SYN, VADER, AT-LSTM, ATAE-LSTM and GCAE, the compared techniques also include:

\begin{itemize}		
	
	\item \textbf{IAN~\cite{MaLZW17}.} Referring to the interactive attention network, it models targets and contexts separately and learn their own representations via interactive learning. By modeling targets and contexts separately, it can pay close attention to the important parts in the target and context;
	
	\item  \textbf{RAM~\cite{ChenSBY17}.} It is a multiple-attention network where the features from multiple attentions are non-linearly combined with a recurrent neural network. It can effectively capture sentiment features separated by a long distance, and is usually more robust against irrelevant information;		
	
	\item \textbf{AOA~\cite{HuangOC18}.} Referring to the attention-over-attention network, it models aspect and sentence in a joint way and benefits from modeling the interaction among word-pairs between sentences and targets;
	
	\item  \textbf{TNet~\cite{LamLSB18}.} It is a target-specific transformation network that employs a CNN layer to extract salient features from the transformed word representations originated from a RNN layer. It avoids using attention for feature extraction so as to alleviate the attended noise.
\end{itemize} 

The rest of this section is organized as follows: Subsection~\ref{sec:setup} describes the experimental setup. Subsection~\ref{sec:comparison} presents the comparative evaluation results. Subsection~\ref{subsec:easy-eval} evaluates the performance of easy instance labeling. Subsection~\ref{sec:sensitivity} evaluates the performance sensitivity of the proposed solution w.r.t various parameters. Finally, Subsection~\ref{sec:scalability} evaluates the scalability of the proposed solution.

\subsection{Experimental Setup} \label{sec:setup}   
In the empirical evaluation, we have used six benchmark datasets in four domains (phone, camera, laptop and restaurant) and two languages (English and Chinese) from the SemEval 2015 task 12 \cite{PontikiGPMA15} and 2016 task 5 \cite{PontikiGPAMAAZQ16}. In all the experiments, we perform 2-class classification to label an aspect polarity as \emph{positive} or \emph{negative}. Note that the datasets of LAP16, RES16, LAP15 and RES15 contain some neutral instances, which are simply ignored in our experiments.
There are no labeled aspect terms in the Chinese datasets of PHO16 and CAM16. Therefore, for ATSA, we only compare GML to its alternatives on the English datasets.


For DNN models, we used the Glove embeddings~\footnote{https://nlp.stanford.edu/projects/glove/} for English data, and the word embeddings from  Baidu~\footnote{http://pan.baidu.com/s/1jIb3yr8} for Chinese data. We employed jieba~\footnote{https://github.com/fxsjy/jieba} to tokenize Chinese sentences. In easy instance labeling and feature extraction for GML, we used the open-source Opinion Lexicon~\footnote{https://www.cs.uic.edu/\url{~}liub/FBS/sentiment-analysis.html} for English data, and the EmotionOntology~\footnote{http://ir.dlut.edu.cn/EmotionOntologyDownload} and BosonNLP~\footnote{https://bosonnlp.com/dev/resource} lexicons for Chinese data. 
For easy instance identification, the scores for the sentiment words in the Chinese lexicons are normalized into the range of [-4, 4], and we use the sentiment words whose scores are at least 1.	In our implementation of GML, the initial weights of word features, similar relational features and opposite relational features are set to 0, 2 and -2 respectively. In the process of scalable gradual inference, if none of the unlabeled instances receives any evidential support from the labeled easier instances, GML employs the existing unsupervised method of LEX-SYN~\cite{Alvarez-LopezJG16} to label its polarity.
%




In the comparative study, for GML and DNN models, we report the average and standard deviation of accuracy over ten runs. 
Our implementation codes of GML have been made open-source available~\footnote{http://www.wowbigdata.com.cn/GML-ALSA/GML-ALSA.html}.


\subsection{Comparative Evaluation}
\label{sec:comparison}

In the comparative study, we set $m$=20, $k$=3, $d_f$=$d_f^*$=0.4, and $d_{f'}$=$d_{f'}^*$=0.1 for GML. Our sensitivity evaluation in Subsection~\ref{sec:sensitivity} has shown that the performance of GML is, to a large extent, insensitive to the parameter setting.


%

\begin{table}
	\footnotesize
	\caption{Accuracy comparison for ACSA on benchmark datasets.}
	\centering
	\begin{tabular}{l c c c c c c c}
		
		\toprule[1.5pt] 
		
		\textbf{Model}
		&  \textbf{PHO16}    & \textbf{CAM16}    & \textbf{LAP16}   
		& \textbf{RES16}  & \textbf{LAP15}   & \textbf{RES15} 
		\\ \hline

		LEX-SYN 
		&  68.43\%   &  78.17\%    & 69.64\%
		& 76.77\% & 75.81\% & 75.31\%
		\\ 
		
		VADER
		& -- 	& --  & 68.31\% & 75.18\%  &74.31\%  & 75.59\%
		\\
		
		\hline
		
		H-LSTM      
		&  (73.30 $\pm$ 0.19) \%       & (78.80$\pm$ 0.60)\%       &  (77.68 $\pm$ 0.65)\%     
		& (81.44$\pm$ 0.39)\%   & (79.03$\pm$ 0.48)\%   &(73.13$\pm$ 1.26) \%
		\\ 
		
		AT-LSTM   
		&  (73.27$\pm$ 1.21)\%      &  \bf (82.49$\pm$ 0.58)\%     &(76.32 $\pm$ 0.74) \%      
		&(83.00$\pm$ 0.43)\%   & (79.03 $\pm$ 1.00)\%  & (76.52 $\pm$ 1.51)\%
		\\
		
		ATAE-LSTM   
		&  (72.40$\pm$ 0.82)\%      &  (81.12$\pm$ 0.70)\%     &(77.90 $\pm$ 1.10) \%      
		&(83.81$\pm$ 1.08)\%   & (79.88 $\pm$ 0.79)\%  & (79.42 $\pm$ 1.07)\%
		\\
		
		GCAE      
		&  \bf (76.94$\pm$ 0.48)\%      &  (82.12$\pm$ 0.53)\%     & \bf (81.94 $\pm$ 0.40) \%      
		& \bf (86.44$\pm$ 0.61)\%   & (82.21 $\pm$ 0.51)\%  & \bf (79.81 $\pm$ 0.70)\%  \\ \hline
		
		GML
		&  (76.14$\pm$ 0.63)\%      &  (81.41$\pm$ 0.25)\%     &(79.84 $\pm$ 0.34) \%      
		&(85.31 $\pm$ 0.06)\%   & \bf (83.94 $\pm$ 0.28)\%  & (78.57 $\pm$ 0.48)\% \\
		
		\bottomrule[1.5pt]	
	\end{tabular}
	\label{table:comparison_ACSA}
\end{table}

%
%
%
%
%
%
%
%
%
%
%
%
%
%
%
%
%
%

\begin{table}
	\footnotesize
	\caption{Accuracy comparison for ATSA on benchmark datasets.}
	\centering
	\begin{tabular}{l c c c c c}
		\toprule[1.5pt] 
		
		\textbf{Model}
		& \textbf{LAP16}   & \textbf{RES16}  & \textbf{LAP15}   & \textbf{RES15} 
		\\ \hline

		LEX-SYN 
		& 67.01\%	& 79.90\%  & 76.55\%  &  75.82\%
		\\ 
		
		VADER
		& 67.22\%	& 79.39\%  &77.11\%  &78.50\%
		\\
		
		\hline
		
		AT-LSTM    
		& (75.49 $\pm$ 1.22) \%      
		& (88.11$\pm$ 0.39)\%   & (80.08 $\pm$ 0.87)\%  &  (76.99 $\pm$ 1.53)\%
		\\
		
		ATAE-LSTM  
		& (76.91 $\pm$ 0.50) \%      
		& (85.44$\pm$ 0.95)\%   & (78.65 $\pm$ 0.64)\%  &  (74.98 $\pm$ 0.98)\%
		\\

		GCAE   
		& \bf (80.21 $\pm$ 0.83)\%      
		& \bf (90.07$\pm$ 0.47)\%   & (80.26 $\pm$ 0.82)\%  &  (78.31 $\pm$ 0.64)\%
		\\ 	\hline
		
		IAN  
		& (78.04 $\pm$ 0.43) \%      
		& (87.50$\pm$ 0.44)\%   & (79.43 $\pm$ 0.81)\%  &  (78.34 $\pm$ 1.02)\%
		\\ 
		
		RAM  
		& (80.21 $\pm$ 1.26) \%      
		& (87.74  $\pm$ 0.45)\%   & (80.49 $\pm$ 0.88)\%  &  (77.61  $\pm$ 1.04)\%
		\\ 
		
		AOA  
		& (78.04 $\pm$ 0.74) \%      
		& (87.80 $\pm$ 0.47)\%   & (81.13 $\pm$ 0.40)\%  &  (78.88 $\pm$ 0.58)\%
		\\ 	
		
		TNet  
		& (79.16 $\pm$ 1.10) \%      
		& (86.99$\pm$ 0.49)\%   & (79.06 $\pm$ 0.79)\%  &  (76.37 $\pm$ 0.89)\%
		\\ 	
		\hline

		GML      
		&  (80.13 $\pm$ 0.31) \%      
		& (85.54 $\pm$ 0.64)\%  &  \bf (82.48 $\pm$ 0.44)\% &  \bf (80.58$\pm$ 0.22)\%   
		\\ \bottomrule[1.5pt]	
	\end{tabular}
	\label{table:comparison_ATSA}
\end{table}

The detailed evaluation results for ACSA and ATSA are presented in Table~\ref{table:comparison_ACSA} and ~\ref{table:comparison_ATSA} respectively. 
Note that VADER can not be directly applied on Chinese data, therefore we only report its performance on English data.   
The best result achieved on each dataset is also highlighted in the table. 
We can see that GML consistently outperforms the unsupervised alternatives, LEX-SYN and VADER (their performance difference is less than 1\% in most cases), by considerable margins on all the test datasets.
For ACSA, the improvement margins on \emph{PHO16},  \emph{RES16} and \emph{LAP15} are around 7-9\%; 
the margins on \emph{LAP16} is even larger at more than 10\%.
For ATSA, it achieves the improvements of more than 10\% on LAP16 and around 5\% on both RES16 and LAP15.
Due to the widely recognized challenge of sentiment analysis, the achieved improvements can be deemed very considerable. \


Furthermore, it can be observed that the performance of GML is highly competitive compared to the supervised DNN techniques. Except GCAE, GML achieves overall better performance than all the other DNN models on both ACSA and ATSA.
For instance, for ACSA, GML beats both AT-LSTM and ATAE-LSTM in performance on five out of totally six datasets.
For ATSA, GML achieves the best performance on two out of totally four datasets; except GCAE, it outperforms all the other DNN model on at least three out of the four datasets. GML even beats GCAE on the ACSA task of \emph{LAP15} and the ATSA tasks of \emph{LAP15} and \emph{RES15}; their performance on the other datasets are close. 
\emph{It is worthy to point out that unlike the DNN models, GML does not use any labeled training data provided in the benchmark.} These experimental results evidently demonstrate the efficacy of GML.


\subsection{Evaluation of Easy Instance Labeling} \label{subsec:easy-eval}

In this subsection, we first evaluate the performance of the proposed technique for identifying easy instances, and then its effect on the performance of GML by comparative study. 
We have compared our proposed technique with VADER, which has been empirically shown to perform slightly better than LEX-SYN.
In our setting,  VADER considers an instance as easy if the absolute value of its sentiment score is more than a threshold (e.g., 0.4 and 0.5). Note that the overall sentiment strength for VADER is a float within the range [-1.0, 1.0].  The detailed evaluation results on the ACSA and ATSA tasks are presented in Table~\ref{table:easy_results}. 
It can be observed that  
\begin{enumerate}
	\item A considerable portion of aspect polarities in the test datasets (varying from 48\% to 60\%) can be identified by our proposed technique as easy, and its accuracy is always high at more than 90\%;
	\item Compared with our proposed technique, VADER with the threshold of 0.4 (i.e., VADER($thres$=0.4)) can identify more easy instances but with considerably lower accuracy; 
	\item Compared with our proposed technique, VADER with the threshold of 0.5 (i.e., VADER($thres$=0.5)) can only identify less easy instances, and its accuracy is also lower in most cases. 	
	
\end{enumerate}

We have also evaluated the effect of different techniques for easy instance labeling on the performance of GML. The detailed evaluation results are presented in Table~\ref{table:easy_compare}, in which GML-vader0.4 (resp. GML-vader0.5) denotes GML that identifies easy instances using VADER with the threshold of 0.4 (resp. 0.5). It can be observed that compared with GML-vader0.4 and GML-vader0.5, GML achieves better performance on seven out of totally eight datasets. Our experimental results have clearly validated the efficay of our proposed technique for easy instance labeling.

\begin{table} 
	\footnotesize
	\caption{Evaluation of easy instance labeling: \emph{Prop} and \emph{Acc} denote the proportion and achieved accuracy of identified easy instances respectively.}
	\centering
	\begin{tabular}{c|c|cccc|cccc}
		\toprule[1.5pt]
		&
		&\multicolumn{4}{c|}{\textbf{ACSA}}
		& \multicolumn{4}{c}{\textbf{ATSA}}
		\\ 
		
		&  & \textbf{LAP16}   & \textbf{RES16}  & \textbf{LAP15}   & \textbf{RES15} & \textbf{LAP16}   & \textbf{RES16}  & \textbf{LAP15}   & \textbf{RES15}
		\\ \hline

		\multirow{2}{*}{\begin{tabular}[l|]{@{}l@{}}VADER\\ ($thres$=0.4) \end{tabular}} 	
		& Prop        & 48.87\%     & 62.34\% 
		& 51.50\%  & 58.21\%  & 50.52\% & 69.54\% & 54.78\% & 62.57\%
		\\ 		
		&Acc    &85.29\%      &90.51\%  
		&87.02\%  & 85.07\%  
		& 86.36\% & 93.69\%  & 88.01\%  &90.80\%
		\\  \hline
		
		\multirow{2}{*}{\begin{tabular}[l|]{@{}l@{}}VADER\\ ($thres$=0.5) \end{tabular}} 	
		& Prop       & 33.82\%     & 48.63\% 
		& 36.98\%  & 46.48\%  & 35.91\% & 56.59\% & 41.65\% & 50.48\%
		\\ 		
		&Acc   &84.65\%      &93.47\%  
		&89.72\%  & 87.24\%  
		& 84.88\% & 95.82\%  & 89.64\%  &92.02\%
		\\	\hline
		
		\multirow{2}{*}{\begin{tabular}[l|]{@{}l@{}}Our\\ approach \end{tabular}} 	
		& Prop       & 48.07\%     & 56.85\% 
		& 55.41\%  & 48.83\%  & 49.06\% & 60.30\% & 56.66\% & 51.44\%
		\\ 		
		&Acc    &92.80\%      &92.64\%  
		&94.59\%  & 92.37\%  
		& 95.32\% & 93.00\%  & 95.36\%  &94.03\%
		\\  
		\bottomrule[1.5pt]	
	\end{tabular}
	\label{table:easy_results}
\end{table}

\begin{table} 	
	\footnotesize
	\caption{Performance comparision between GML-vader0.4, GML-vade0.5 and GML.}
	\centering
	\begin{tabular}{l|cccc|cccc}
		\toprule[1.5pt] 
		&\multicolumn{4}{c|}{\textbf{ACSA}}
		& \multicolumn{4}{c}{\textbf{ATSA}}
		\\

		& \textbf{LAP16}   & \textbf{RES16}  & \textbf{LAP15}   & \textbf{RES15} & \textbf{LAP16}   & \textbf{RES16}  & \textbf{LAP15}   & \textbf{RES15}
		\\ \hline
		
		GML-vader0.4     & 76.35\%     & 83.29\% 
		& 79.33\%  & 75.83\%  & 75.37\% & \bf 87.13\% & 76.59\% & 79.04\%
		\\ 		
		GML-vader0.5   & 74.35\%      &82.86\%  
		&75.46\%  & 76.55\%  
		& 71.69\% & 86.05\%  & 74.71\%  &76.47\%
		\\ 
		
		GML   
		& \bf 79.84\% & \bf 85.31\%  & \bf 83.94\%  & \bf 78.57\%     
		& \bf 80.13\%      &85.54\%  
		& \bf 82.48\%  & \bf 80.58\%  		
		\\ 		
		\bottomrule[1.5pt]	
	\end{tabular}
	\label{table:easy_compare}
\end{table}

\subsection{Sensitivity Evaluation} \label{sec:sensitivity}

\begin{table}[!h]
	\footnotesize
	\centering
	\caption{Sensitivity evaluation over ACSA tasks.}
	\label{table:sensitivity_ACSA}
	\begin{tabular}{c | c | c | c | c | c | c | c c}
		\toprule[1.5pt] 
		\multirow{7}{*}{\begin{tabular}[l]{@{}l@{}}w.r.t  $\bm{m}$ \\ ($k=3$) \end{tabular}}	
		&\multicolumn{1}{c|}{} & \multicolumn{1}{c|}{\textbf{PHO16}} & \multicolumn{1}{c|}{\textbf{CAM16}}  & \multicolumn{1}{c|}{{\textbf{LAP16}}}  & \multicolumn{1}{c}{\textbf{RES16}}  
		& \multicolumn{1}{c|}{\textbf{LAP15}}  
		& \multicolumn{1}{c}{\textbf{RES15}} \\ 
		\cmidrule{1-8}
		& $m=10$                  
		& 75.91\%                      &81.54\%                       
		& 79.95\%                      & 83.93\%                      
		& 84.24\%	                   & 79.39\% \\
		
		& $m=20$                
		
		& 76.14\%     & 81.41\%       & 79.84\%  
		& 85.31\%	   &  83.94\%   & 78.57\%  \\ 
		
		& $m=30$                   
		& 76.26\%                      & 81.25\%                      
		& 80.08\%                       & 84.79\%                       
		& 83.94\%	&79.47\%\\
		
		& $m=40$                 
		& 75.61\%                      & 80.96\%                       
		& 80.08\%                       & 84.76\%                       
		& 84.03\%	&79.11\%\\ 
		\hline
		
		\multirow{4}{*}{\begin{tabular}[l]{@{}l@{}}w.r.t  $\bm{k}$ \\ ($m=20$) \end{tabular}}
		& $k=1$                  
		& 77.28\%                      &81.58\%                       
		& 79.95\%                      & 84.99\%                      
		& 83.06\%	                   & 79.81\% \\
		
		& $k=3$                 
		& 76.14\%     & 81.41\%       & 79.84\%  
		& 85.31\%	   &  83.94\%   & 78.57\%  \\ 
		
		& $k=5$                   
		& 75.69\%                      & 80.91\%                      
		& 77.90\%                       & 85.11\%                       
		& 82.97\%	&79.08\%\\
		
		& $k=7$                 
		& 75.92\%                      & 80.54\%                       
		& 80.05\%                       & 85.08\%                       
		& 83.16\%	&78.69\%\\ 
		\bottomrule[1.5pt]	
	\end{tabular}
	\begin{tabular}{c | c | c | c | c | c | c | c c}
		\toprule[1.5pt]	
		\multirow{9}{*}{\begin{tabular}[]{@{}l@{}} {}w.r.t  $\bm{d_{f'}}$ \\ and $\bm{d_{f}}$ \\ ($m=20$ \\ \ $k=3$) \end{tabular}}		
		&\multicolumn{1}{c|}{$d_{f'}$ \ $d_{f}$} & \multicolumn{1}{c|}{\textbf{PHO16}} & \multicolumn{1}{c|}{\textbf{CAM16}}  & \multicolumn{1}{c|}{{\textbf{LAP16}}}  & \multicolumn{1}{c}{\textbf{RES16}}  
		& \multicolumn{1}{c|}{\textbf{LAP15}}  
		& \multicolumn{1}{c}{\textbf{RES15}} \\ 
		\cmidrule{2-8}
		& ${0.1 \ \  0.2}$
		& 76.60\%                      &81.25\%                       
		& 79.81\%                      & 82.91\%                      
		& 84.22\%	                   & 79.64\% \\
		
		&${0.1 \ \  0.3}$                
		& 76.67\%                      & 80.79\% 
		&79.76\%                       & 84.53\%                      
		& 83.96\%                       &78.64\%  \\
		
		&${0.1 \ \  0.4}$ 	
		
		& 76.14\%     & 81.41\%       & 79.84\%  
		& 85.31\%	   &  83.94\%   & 78.57\%  \\ 
		
		&${0.2 \ \  0.3}$              
		& 76.45\%                      & 80.91\%                       
		& 79.73\%                       & 84.82\%                       
		& 84.08\%	&78.97\%\\ 
		
		&${0.2 \ \  0.4}$               
		& 76.41\%                      & 81.08\%                       
		& 79.89\%                       & 84.59\%                       
		& 84.01\%	&78.72\%\\ 
		
		&${0.2 \ \  0.5}$               
		& 76.26\%                      & 81.41\%                       
		& 79.81\%                       & 85.05\%                       
		& 84.12\%	&79.69\%\\ 
		
		&${0.3 \ \  0.4}$               
		& 75.95\%                      & 80.96\%                       
		& 79.79\%                       & 84.62\%                       
		& 83.99\%	&78.92\%\\ 
		
		&${0.3 \ \  0.5}$               
		& 76.22\%                      & 80.91\%                       
		& 80.16\%                       & 85.25\%                       
		& 83.64\%	&78.58\%\\ 	
		\bottomrule[1.5pt]	
	\end{tabular}
\end{table}

In the sensitivity evaluation, we first vary the values of the parameters $m$ and $k$ as shown in Algorithm~\ref{alg:gradualinference}, which respectively denote the number of candidate variables selected for approximate entropy ranking and the number of candidate variables selected for factor graph inference. We set $m$=10, 20, 30, 40, and $k$=1, 3, 5, 7. We then vary the values of the parameters, $d_{f}$, $d_{f'}$, $d_f^*$ and $d_{f'}^*$, which denote the degree of uncertainty of word and relational features as shown in Eq.~\ref{eq:word-mass}, ~\ref{eq:relation-mass}, ~\ref{eq:word-rank} and ~\ref{eq:relation-rank}. We set $d_f$=$d_f^*$, $d_{f'}=d_{f'}^*$, and $d_{f'}<d_{f}<=0.5$.  While evaluating GML's sensitivity to a particular parameter, we fix any other parameter to the same value. 

The detailed evaluation results on ACSA are presented in Table~\ref{table:sensitivity_ACSA}. Since the standard deviations of accuracy are very similar under different parameter setting, we only report the averages of accuracy in the table. The evaluation results on ATSA are similar, thus omitted here. It can be observed that the performance of GML only fluctuate slightly ($\leq 1\%$ in most cases) with different parameter settings. It is noteworthy that the performance of GML does not fluctuate much with various values of $m$ and $k$. Since most of GML's runtime is spent on factor graph inference, reducing the value of $k$ can effectively improve efficiency. Our experiments show that even with $k$ taking the minimal value of $1$, the performance of GML only changes marginally. We also have the similar observation on the parameter setting of $d_{f}$ and $d_{f'}$. Various value combinations of $d_{f}$ and $d_{f'}$ can only result in very marginal performance fluctuations. Our experimental results clearly show that the performance of GML is, to a large extent, insensitive to the parameter settings. They bode well for its applicability in real scenarios.  

\subsection{Scalability Evaluation} \label{sec:scalability}

\begin{figure}
	\centering
	\subfigure[]{
		\label{fig:subfig:a} 
		\includegraphics[width=2.8in]{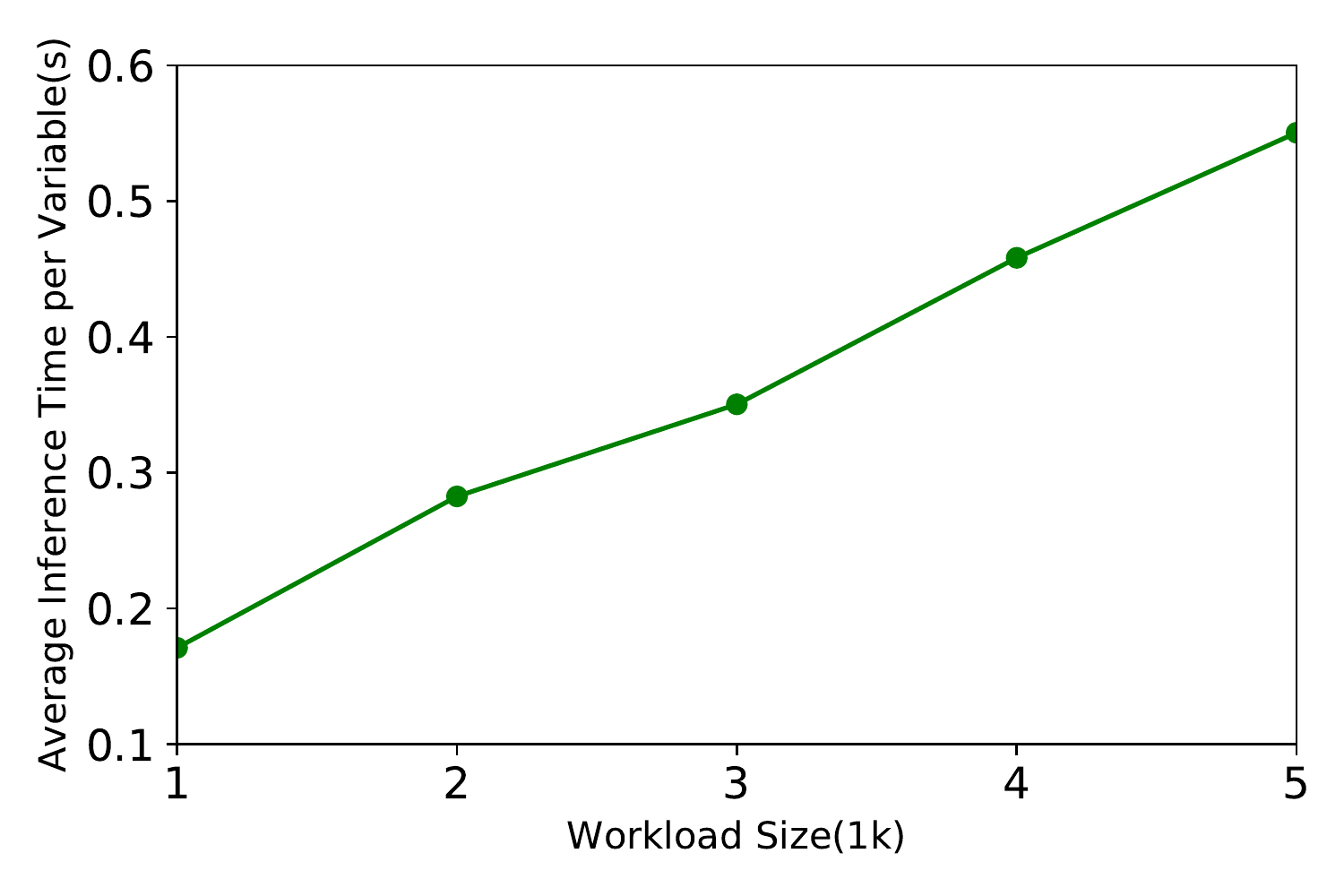}}
	\subfigure[]{
		\label{fig:subfig:b} 
		\includegraphics[width=2.8in]{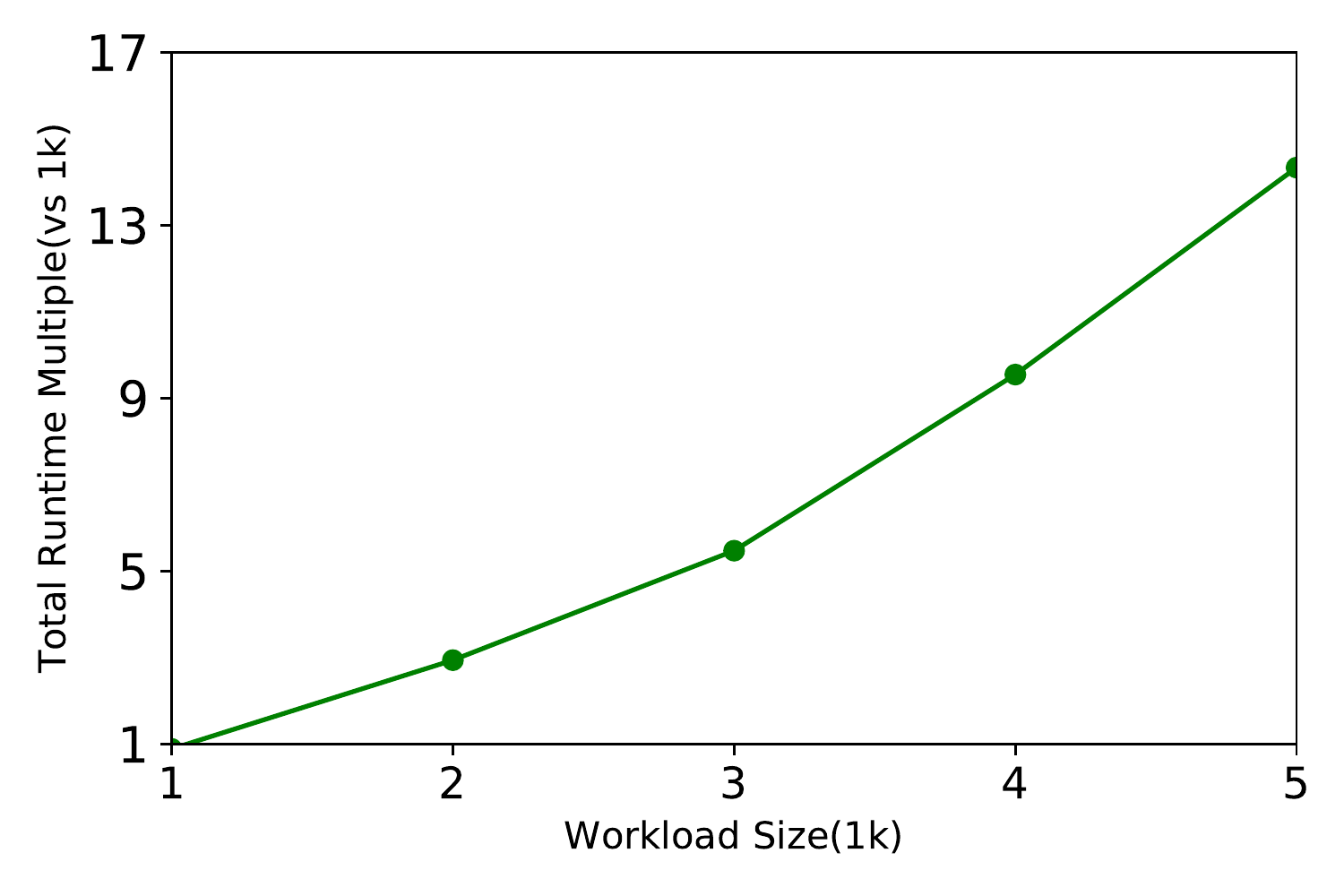}}
	\caption{Scalability evaluation.}
	\label{fig:scalability} 
\end{figure}

In this section, we evaluate the scalability of the proposed scalable approach for GML. We have generated the restaurant workloads with different sizes by retrieving the reviews from Yelp. The workload size varies from 1000 to 5000.  All the algorithmic parameters are set to the same values for different workloads. In the experiments, we set $m$=20, $k$=3, $d_f$=$d_f^*$=0.4, and $d_{f'}$=$d_{f'}^*$=0.1. The detailed evaluation results on runtime are presented in Figure~\ref{fig:scalability}. We have observed that most of the runtime is spent on factor graph inference. Even though the total number of features in the factor graph is large, the number of features an instance has is usually quite limited. As a result, the size of the subgraph constructed for scalable factor inference on an unlabeled variable generally increases linearly with workload size. Accordingly, the average computational cost of the scalable GML spent on each unlabeled variable increases nearly linearly with workload size. Therefore, as shown in Figure~\ref{fig:scalability}, the proposed scalable approach scales well with workload size.




\section{Conclusion} \label{sec:conclusion}

In this paper, we have proposed a technical solution
for the task of ALSA based on the recently proposed paradigm of gradual machine learning. It begins with some easy instances in an ALSA task, and then gradually labels the more challenging instances based on iterative factor graph inference without requiring any human intervention.
Our empirical study on the benchmark datasets has validated the efficacy of the proposed solution.

Our research on gradual machine learning is an ongoing effort. Future work can be pursued on several fronts. Even though GML has been proposed as unsupervised learning approach, human work can be potentially integrated into its process for improved performance. An interesting open challenge is then how to effectively improve the performance of gradual machine learning for ALSA with the minimal effort of human intervention, which include but are not limited to manually labeling some instances. It is also interesting to develop the solution of gradual machine learning for the challenging classification tasks other than entity resolution and sentiment analysis.


\bibliography{bibfile}

\end{document}